\documentclass[sigconf,natbib=true,anonymous=false]{acmart}
\AtBeginDocument{%
  \providecommand\BibTeX{{%
    \normalfont B\kern-0.5em{\scshape i\kern-0.25em b}\kern-0.8em\TeX}}}

\setcopyright{acmlicensed}
\copyrightyear{2018}
\acmYear{2018}
\acmDOI{XXXXXXX.XXXXXXX}

\acmConference[Conference acronym 'XX]{Make sure to enter the correct
  conference title from your rights confirmation emai}{June 03--05,
  2018}{Woodstock, NY}
\acmBooktitle{Woodstock '18: ACM Symposium on Neural Gaze Detection,
 June 03--05, 2018, Woodstock, NY} 
\acmISBN{978-1-4503-XXXX-X/18/06}

\usepackage{balance}

\usepackage{bbm}
\usepackage{subcaption}
\usepackage{multirow}
\usepackage{enumitem}
\usepackage{bm} 
\usepackage{amsmath,amsfonts}
\usepackage{graphicx}
\usepackage{graphics}
\usepackage{textcomp}
\usepackage{xcolor}
\usepackage{hyperref}
\usepackage[linesnumbered,ruled,procnumbered,vlined]{algorithm2e}

\SetCommentSty{mycommfont}
\SetKwInput{KwInput}{Input}                
\SetKwInput{KwOutput}{Output}              
\SetKwInput{Parameters}{Parameters}              
\newcommand{\modelname}{IM-GNB}
\newcommand{\LL}{J}

\newcommand{\eat}[1]{}
\copyrightyear{2024}
\acmYear{2024}
\setcopyright{rightsretained}
\acmConference[KDD '24]{Proceedings of the 30th ACM SIGKDD Conference on
Knowledge Discovery and Data Mining}{August 25--29, 2024}{Barcelona, Spain}
\acmBooktitle{Proceedings of the 30th ACM SIGKDD Conference on Knowledge
Discovery and Data Mining (KDD '24), August 25--29, 2024, Barcelona,
Spain}
\acmDOI{10.1145/3637528.3671983}
\acmISBN{979-8-4007-0490-1/24/08}

\makeatletter
\def\author@bx@sep{1pt}
\makeatother

\begin{document}

\title{Influence Maximization via Graph Neural Bandits}

\author{Yuting Feng}
\affiliation{%
  \institution{Department of Electrical and  Computer Engineering (ECE)} 
  \city{National University of Singapore}
  \country{}
}
\email{yt.f@nus.edu.sg}

\author{Vincent Y.~F.~Tan}
\affiliation{%
  \institution{Department of Mathematics,\\ Department of ECE, }
  \streetaddress{}
  \city{National University of Singapore}
  \country{}}
\email{vtan@nus.edu.sg}

\author{Bogdan Cautis}
\affiliation{%
  \institution{University of Paris-Saclay,\\ CNRS LISN}
  \country{}
}
\email{bogdan.cautis@universite-paris-saclay.fr}






\renewcommand{\shortauthors}{Feng, Tan, and Cautis}

\begin{abstract}

We consider a ubiquitous scenario in the study of Influence Maximization (IM), in which there is limited knowledge about the topology of the diffusion network. We set the IM problem in a multi-round diffusion campaign, aiming to maximize the number of distinct users that are influenced. Leveraging the capability of bandit algorithms to effectively balance the objectives of exploration and exploitation, as well as the expressivity of neural networks, our study explores the application of neural bandit algorithms to the IM problem. We propose the framework {\sc IM-GNB} (Influence Maximization with Graph Neural Bandits), where we provide an estimate of the users' probabilities of being influenced by influencers (also known as diffusion seeds). This initial estimate forms the basis for constructing both an exploitation graph and an exploration one. Subsequently, {\sc IM-GNB} handles the exploration-exploitation tradeoff, by selecting seed nodes in real-time using Graph Convolutional Networks (GCN), in which the pre-estimated graphs are employed to refine the influencers' estimated rewards in each contextual setting. Through extensive experiments on two large real-world datasets, we demonstrate the effectiveness of {\sc IM-GNB} compared with other baseline methods, significantly improving the spread outcome of such diffusion campaigns, when the underlying network is unknown.
\vspace{-3mm}
\end{abstract}




\maketitle
\vspace{-1mm}
\section{Introduction}
\label{sec:intro}
Motivated by the rise of ``influencer marketing'' in social media advertising, a class of algorithmic problems termed \emph{Influence Maximization} (IM) has emerged, starting with the pioneering work of \cite{DBLP:conf/kdd/DomingosR01,kempe2003maximizing}. These algorithms aim to identify the most influential nodes within a diffusion network for initiating the spread of specific information, thereby maximizing its reach. In many ways, this research directly mirrors the increasingly prevalent and successful marketing strategy of targeting key individuals (influencers).

The objective of IM is typically formulated by maximizing the expected \emph{spread} under a stochastic diffusion model, which characterizes the information dissemination process. The work of \cite{kempe2003maximizing}  laid the foundations for the IM literature, by introducing two prominent models: Linear Threshold (LT) and Independent Cascade (IC). These models, widely adopted in subsequent research, represent diffusion networks as probabilistic graphs, where the edges are weighted by  probabilities of information transmission. 

Selecting the seed nodes maximizing the expected spread is NP-hard under   common diffusion models \cite{kempe2003maximizing}. Despite the development of approximate algorithms, exploiting the monotonicity and submodularity of the spread, scaling IM algorithms to large networks remains challenging. Acquiring meaningful influence probabilities is equally challenging, as learning them from past information cascades (e.g., as in \cite{GomezRodriguezS12,DuSGZ13}) can be data-intensive and thus impractical. Moreover, the applicability of such models is limited in scenarios where historical cascades are not available.

In the face of these challenges, since even the most efficient IM algorithms such as \cite{ DBLP:journals/pvldb/HuangWBXL17,DBLP:conf/sigmod/TangSX15} rely on assumptions and parameters that often fail to capture the complex reality of how information spreads online, a change in research direction has been followed recently. It consists of approaches that 
neither rely on pre-defined diffusion models nor require upfront knowledge of the diffusion network. Instead, these \emph{online} methods, such as \cite{pmlr-v70-vaswani17a, lagree2018algorithms, iacob2022contextual}, \emph{learn} to spread on the fly. More precisely, they involve a \emph{sequential} learning agent that actively gathers information through a \emph{multi-round influence campaign}. In each round, the agent selects so-called {\em seed nodes}, observes the resulting information spread, and uses this feedback to make better choices in subsequent rounds, with the campaign's total reward being the objective that is to be optimized. Such a learning framework leads naturally to a policy that balances  {\em exploring} unknown aspects (i.e., the diffusion dynamics) with {\em exploiting} known and successful choices (i.e., the high-performing spread seed nodes), using  \emph{multi-armed bandits} \cite{lattimore2020bandit}.

We consider in this paper such an an online IM scenario with limited network information. Specifically, the diffusion graph is largely unknown, except for a set of predefined \emph{influencers}, representing the potential seeds for information dissemination at each round of a multi-round diffusion campaign. Additionally, we incorporate \emph{contextual features} of both influencers and the information being diffused. Regarding the latter, the rationale is that within a campaign aiming to maximize the reach of a specific message, its framing and presentation can significantly impact its spread. For instance, a political campaign may use various formats like news articles, opinion pieces, data visualizations, or multimedia content, each leading to distinct diffusion patterns.


We leverage such contextual information through the formal framework of Contextual Multi-Armed Bandits (CMABs) \cite{lattimore2020bandit}. Furthermore, recognizing that significant correlations between the features of the \emph{basic (to-be-influenced) users} may exist, albeit unknown to the agent, and that by implication their activation probabilities may be correlated, we enhance the learning framework with mechanisms by which \emph{each activation} can provide useful information about \emph{neighbouring users} in the network as well, allowing to refine the agent’s predictions. We achieve this by adapting to our IM problem setting the Graph Neural Bandits (GNB) framework of \cite{qi2023graph} (a bandit algorithm for recommender systems). Correlation graphs are constructed based on the similarity of users to be influenced by the same influencer, and GNBs are then employed to handle the challenges associated with graph-based bandit algorithm. In doing so, our work is the first to leverage the implicit relationships that may exist between basic users in the unknown diffusion medium. In essence,  we dynamically model these relationships based on the observed campaign feedback, and we use them as input for a graph neural network (GNN)-based learning algorithm guiding the seed selection process at each round, optimizing choices under the exploit-explore paradigm.

\noindent \textbf{Overview of our IM scenario.} As usual in IM scenarios, we run campaigns under budget constraints (limited seedings and rounds), with the goal to maximize the number of \emph{distinct users activated}, starting from known influencers. The learning agent chooses seeds sequentially, i.e., at each round, with potential \emph{re-seeding}, and feedback consists solely of the activated nodes after each round, without additional details on the triggering causes. The feedback is used to refine estimates of influencer potential, guiding future seeding choices. Aligning with the overall objective, each round's reward is the number of newly activated users, and the campaign aims to maximize the cumulative reward across rounds. In this scenario, we mimic real-world influencer marketing, where access is limited to a few influencers, feedback is restricted to user actions (like purchases or subscriptions), and the goal is to reach as many \emph{unique} users as possible.

\noindent \textbf{Our contributions.}  We detail our contributions in the following:
 \vspace{-1mm}
\begin{itemize}[leftmargin=*]
    \item  By introducing the IM-GNB framework, we 
    connect GNBs and the IM problem. This integration is non-trivial due to the inherent challenges of learning from graph-structured data and making sequential decisions under uncertain environments in the context of diffusion campaigns.
    \item We tackle the challenge of balancing exploration and exploitation in dynamic influence propagation by incorporating contextual bandits into the IM-GNB framework. This enables us to effectively explore the potential rewards while exploiting available information, resulting in enhanced influence spread in real-world scenarios.
    \item We construct user-user correlation graphs for exploitation and exploration purposes, capturing intricate interactions among users and influencers in each round of the diffusion campaign. This graph-based approach is scalable to various network settings, even without prior knowledge of the network's topology structure.
    \item We develop a novel algorithm that optimally selects seed nodes in real-time, with contextual bandits integrated with GNNs to refine the reward estimates in each contextual setting. Through extensive experiments, we show that our algorithm outperforms baseline methods, highlighting the utility of GNBs as a principled approach to optimize influence campaigns in uncertain environments.
\end{itemize}


\section{Related Work}
\label{sec:related}
\vspace{-1mm}
\noindent \textbf{Influence Maximization}
 (IM) addresses the challenge of identifying a set of  seeds (influencers) within a social network to maximize information spread. Researchers first explored this problem in \cite{DBLP:conf/kdd/DomingosR01}. Later, \cite{kempe2003maximizing} provided a clear formulation of the problem, including how influence spreads through stochastic models like Independent Cascade (IC) and Linear Threshold (LT). They also described the important properties of the spread objective, its approximation guarantees and hardness results. Since then, such stochastic models have become widely adopted in the literature, and most works focused on finding approximate solutions that can be computed efficiently. A key breakthrough was the concept of reverse influence sampling, introduced in
 \cite{borgs2014maximizing} and  made practical in \cite{TXS14,DBLP:conf/sigmod/TangSX15,nguyen2016stop}. Diffusion model-based IM approaches rely on diffusion graphs where the edges are labeled by weights (spread probabilities). In empirical evaluations, these weights may be data-based \cite{goyal2010learning,goyal2011data} (computed from diffusion cascades), degree based,  or simply assumed random.  Some recent studies~\cite{feng2018inf2vec,panagopoulos2020multi} employ representation learning to infer influence probabilities from ground-truth diffusion cascades, 
 a resource that may not be readily available in many application scenarios.  (See the recent survey~\cite{10.1145/3604559} for a review of the IM literature.) 
 \eat{
 for numerous datasets~\cite{DBLP:conf/kdd/FengPCV23}. 
 }
 \eat{
 The pioneering work~\cite{kempe2003maximizing} introduced two diffusion models, Linear Threshold (LT) and Independent Cascade (IC), to model the stochastic diffusion process for online information spread. These have become widely adopted in the subsequent literature, as shown in the survey~\cite{li2018influence}. 
 However, diffusion model-based IM approaches suffer from oversimplifying assumptions by utilizing random or uniform inﬂuence parameters and these models depend on diffusion graphs where the edges are assigned weights based on spread probabilities. Some model-independent studies~\cite{goyal2010learning,goyal2011data,feng2018inf2vec,panagopoulos2020multi} employ representation learning to infer influence probabilities from ground truth diffusion cascades, contributing to data-intensive models. These approaches specifically require extensive data on historical diffusion cascades, a resource not readily available for numerous datasets~\cite{DBLP:conf/kdd/FengPCV23}. 
 }

\noindent \textbf{Bandits for Influence Maximization}
By virtue of their versatility and sequential nature, bandit algorithms are apt to be used in  IM problems, especially in uncertain diffusion environments with which a learning agent may interact repeatedly~\cite{vaswani2015influence,sun2018multi,wu2019factorization,iacob2022contextual}. A multi-round, sequential setting allows to spread information and gather feedback, striking a balance between influencing / activating nodes in each round and learning influence parameters for uncertain or unexplored network facets. This strategy closely mirrors real-world influencer marketing scenarios, in which campaigns often unfold over time. \cite{vaswani2015influence} is one of the earliest works that map an IM problem formulation to a combinatorial multi-armed bandit (CMAB) paradigm, where diffusions are assumed to follow the IC model. IMLinUCB~\cite{wen2017online} learns the optimal influencers dynamically, while repeatedly interacting with a network under the IC assumption as well. Vaswani et al.~\cite{pmlr-v70-vaswani17a} introduces a diffusion model-agnostic framework, based on a pairwise-influence semi-bandit feedback model and the LinUCB-based algorithm, addressing scenarios involving new marketers that exploit existing networks. Since the aforementioned approaches leverage a given diffusion graph topology, the inherent difficulty of obtaining such data limits their practical interest. 


Operating in highly uncertain diffusion scenarios that (i) make no assumption on the diffusion model and (ii) lack knowledge of the diffusion topology and historical activations (cascades), \cite{lagree2018algorithms} proposes FAT-GT-UCB, where a Good--Turing estimator is used to capture the utility (called remaining potential) of an influencer, throughout the multiple rounds of a diffusion campaign. They also consider a \emph{fatigue} effect for influencers, since these may be are repeatedly chosen in the sequential rounds. GLM-GT-UCB~\cite{iacob2022contextual} considers the same setting as~\cite{lagree2018algorithms}, while exploiting contextual information (e.g., features pertaining to influencers or the information being conveyed). Our work shares a similar setting, where  the network topology is unknown and no assumptions are made about the diffusion model. In a multi-round diffusion campaign, we select at each round diffusion seeds, \emph{without} factoring in influencer fatigue.


\noindent \textbf{Bandits with deep learning}
Early works~\cite{abe2003reinforcement,dani2008stochastic,rusmevichientong2010linearly} in the contextual bandit literature focused on linear models, assuming the expected reward at each round is linear in the feature vector. This assumption, however, often fails to hold in practice, prompting exploration into nonlinear or nonparametric 
contextual bandits~\cite{bubeck2011x,valko2013finite}. However, these more complex models impose  restrictive assumptions on the reward function, such as Lipschitz continuity~\cite{bubeck2011x}, or a reward function from a reproducing kernel Hilbert space (RKHS)~\cite{valko2013finite}.

To overcome these limitations, several recent studies~\cite{riquelme2018deep,zahavy2019deep,zhou2020neural} leverage the expressivity of deep neural networks (DNNs) to incorporate nonlinear models, which require less domain knowledge. The works of~\cite{riquelme2018deep,zahavy2019deep} employ DNNs for effective context transformation with a linear exploration policy, showing notable empirical success despite the absence of regret guarantees. The work of~\cite{zhou2020neural} introduces NeuralUCB, a provably efficient neural contextual bandit algorithm using  DNN-based random feature mappings to construct the UCB, with a near-optimal regret guarantee. The construct of the UCB is based on the past gradient of the exploitation function. The work of~\cite{zhang2020neural} assigns a    normal distribution as the distribution of the reward of each arm, similarly to the deviation computed on the gradient of the estimation function. Similar to some other studies, EE-Net~\cite{ban2021ee} has an exploitation network to estimate rewards for each arm. It additionally builds an exploration network to predict the potential gain for each arm, relative to the current estimate, where the input of the exploration network are the previous gradients of the exploitation function. The work of Qi {\em et al.}~\cite{qi2023graph} employs contextual neural bandits in recommender systems, to build a graph neural bandit framework where each arm is induced with an exploitation graph and an exploration one, with the weights of edges representing users' correlations regarding the exploitation and exploration performed. The effectiveness of~\cite{qi2023graph} in the recommendation setting serves as our initial motivation for leveraging its neural bandits framework in our IM problem. Given the similarities in predicting user preferences (user-item in recommender systems or user-influencer susceptibility in IM), we exploit a graph neural contextual bandit algorithm  to maximize the influence spread in multi-round diffusion campaigns.

\vspace{-1mm}
\section{Problem Formulation}
\label{sec:prob}

We formulate the Influence Maximization (IM) problem with a discrete-time diffusion model~\cite{kempe2003maximizing}, adopting a combinatorial multi-armed bandit paradigm to estimate the influence spread. 

\noindent \textbf{IM Problem} Within the context of information scenarios characterized by stochastic or epidemic information diffusion phenomena, particularly on social media, the information spread is initiated by seed users (influencers) and amplified through sharing and retweeting via user interactions. For a campaign of information spread consisting of $T$ rounds (trials), we select the influencers at each round to maximize the overall information spread.

 
We are given a known base set of influencers $K=\{k_i\}_{i=1}^n$ as seeds, a budget of $T$ rounds (trials). At each round $t \in \{ 1,2 \ldots, T\}$, the environment provides us with the message $C_t$ to diffuse, and there are $L \in \{1,2,\dots,n\} $ seeds to be activated initially. With $I_t$ (which has cardinality $|I_t|=L$) the set of activated seeds, $S(I_t, C_t)$ is the round's spread (all activated users) starting from the chosen seed set $I_t$. Our objective is to maximize the cumulative and \emph{distinct} spread of the $T$ rounds, i.e., find
\begin{equation}
    \underset{I_t\subseteq K,|I_t|=L,\forall1\le t \le T}{\mathrm{argmax}}\mathbb{E} \bigg[\Big|\bigcup_{1\le t \le T}S(I_t, C_t)\Big|\bigg].
\end{equation}
\textbf{Adaptation to the bandit setting} To adapt the IM problem to a contextual bandit setting, the set of influencers $K$ can be considered the set of arms to be pulled in $T$ rounds. At each round $t$, with the provided message $C_t$ as the context, the set of arms $I_t = \{k_i\}_{i=1}^{L}$ is chosen. For each chosen arm $k_i$, $A_{k_i}$ is the set of \emph{basic users} activated or influenced by seed (arm) $k_i$. For each basic user $u$, let $c_u^t$ denote the total number of times it has been influenced or activated until round $t$. With the 
set of activated users (influence spread) as the node semi-bandit feedback, the reward is the number of new activations~\cite{iacob2022contextual} as
\begin{equation}
\label{eq:true_spread}
    R_t = \sum_{u\in{\bigcup_{k_i\in{I_t}}A_{k_i}}}\mathbbm{1}\{c_u^t>0\} - R_{t-1};\qquad R_0=0,
\end{equation}
Note that  distinct activations are used for the cumulative reward, i.e., a given user will be counted only {\em once} in the total reward, even if it has been influenced several times.


\noindent \textbf{Modeling with graph bandits} We are mainly motivated by  application scenarios in social media (e.g., information campaigns for elections, online advertising, public awareness campaigns, crisis information diffusion, etc.), where users may exhibit similar preferences and influence susceptibility for certain diffusion topics (e.g., sharing the same political views) initiated by certain influencers (arms), while they may react differently and be more susceptible to other influencers for other topics (e.g., entertainment or sports).

Thus, instead of representing the social graph uniformly, in the bandit setting, we allow each arm $k_i$ at each round $t$ to induce a  distinct {\em graph} $G_{i,t}({\mathcal{U}}, E, W_{i,t})$ to represent   user connectivity. With $\bm{k}_i$ the $d_1$-dimensional feature vector of arm $k_i$ and $\bm{C}_t$ the $d_2$-dimensional context vector, the \textbf{expected reward}\footnote{Note that this expected reward $r_{i,t}$ is assessing the distinct activations  by arm $k_i$ at round $t$, in alignment with the reward brought by each arm in $I_t$, as defined in Eq.~\eqref{eq:true_spread}.} 
at each round $t\in[T]$ brought by arm $k_i$ is defined as
\begin{equation}
\label{eq:reward_total}
    r_{i,t} = f(\bm{k}_{i},\bm{C}_t,G_{i,t}).
\end{equation}
In $G_{i,t}$, each user $u\in\mathcal{U}=\{1,2,\ldots,m\}$ corresponds to a node,   $E$ is the set of edges connecting users, and $W_{i,t} = \{w_{i,t}(u,u'): u,u'\in\mathcal{U}\}$ is the set of weights corresponding to each edge $e\in E$. Modeling real applications, we assume that the weights of the edges connecting  nodes in $G_{i,t}$ represent  users' similarity w.r.t. the same influencer (arm $k_{i}$), i.e., the probability to be similarly influenced by arm $k_{i}$ in round $t$, which is defined as
\begin{equation}
\label{eq: weights}
    w_{i,t}(u,u') = \Phi^{(1)}\Big(\mathbb{E}\big[p_{i,u}^{t}|\bm{k}_i,\bm{C}_t\big],\mathbb{E}\big[p_{i,u'}^{t}|\bm{k}_i,\bm{C}_t\big]\Big),
\end{equation}
where $p_{i,u}^t = h_u(\bm{k}_i,\bm{C}_t) \in [0,1]$ 
the \textbf{expected diffusion probability} between  influencer (arm $k_i$) and user $u$ under the context   $\bm{C}_t$, and $\Phi^{(1)}: \mathbb{R} \times \mathbb{R} \rightarrow \mathbb{R}$ maps the expected diffusion probability of users w.r.t.\ influencer $k_i$ to the   weights among users in~$G_{i,t}$. 


However, the similarity graph $G_{i,t}$ and the function $h_u$ are  unknown   in our problem setting. Thus we propose an estimate graph $G^{(1)}_{i,t} = ({\mathcal{U}}, E, W^{(1)}_{i,t})$ to approximate $G_{i,t}$ by exploiting the current observations. This is  known  as the \emph{exploitation graph}. We also consider a pre-defined hypothesis function $h_u^{(1)}(\bm{k}_i,\bm{C}_t)$ to approximate the expected diffusion probability $p_{i,u}^t$,
so as to estimate $W_{i,t}$ with $W^{(1)}_{i,t}$ in the graph $G^{(1)}_{i,t}$. With the pre-estimated graph $G^{(1)}_{i,t}$, the estimate reward of arm $k_i$ across all users is then expressed as
\begin{equation}
    \hat{r}_{i,t} = f^{(1)}\big(\bm{k}_{i},\bm{C}_t,G^{(1)}_{i,t}\big).
    \vspace{-1mm}
\end{equation}
To quantify the estimation gap (uncertainty of estimation) between $G^{(1)}_{i,t}$ and $G_{i,t}$ (or to measure the potential gain on the estimated diffusion probability for each user-influencer pair), we also propose an \emph{exploration graph}, denoted by $G^{(2)}_{i,t} = ({\mathcal{U}}, E, W^{(2)}_{i,t})$,  where analogously the weights among users $w^{(2)}_{i,t}(u,u')\in W^{(2)}_{i,t}$ indicate users' correlations w.r.t. potential gains, expressed as
\vspace{-1mm}
\begin{align}
     &w^{(2)}_{i,t}(u,u')   \nonumber\\
    &\;\;= \Phi^{(2)}\big(h_u(\bm{k}_i,\bm{C}_t)- 
    h_u^{(1)}(\bm{k}_i,\bm{C}_t),h_{u'}(\bm{k}_i,\bm{C}_t) - h_{u'}^{(1)}(\bm{k}_i,\bm{C}_t)\big) \!\label{eq:weight2}
    \vspace{-1mm}
\end{align}
for some function $\Phi^{(2)}:\mathbb{R} \times \mathbb{R} \rightarrow \mathbb{R}$ which is similar to $\Phi^{(1)}$. 

With the exploration graph $G^{(2)}_{i,t}$, the potential gain of arm $k_i$ across all the users is defined as $\hat{b}_{i,t} = f^{(2)}(\bm{k}_{i},\bm{C}_t,G^{(2)}_{i,t})$.  At each round~$t$, the arm set $I_t$ is selected as $\arg\max_{I_t\subset K:|I_t|=L}(\hat{r}_{i,t}+\hat{b}_{i,t})$. This maximizes the overall influence spread in the campaign.

The details of the constructions of the exploitation and exploration graphs are given in Sec.~\ref{sec:method} below.










\vspace{-1mm}
\section{Proposed framework}
\label{sec:method}
\begin{figure*}
  \centering
  \includegraphics[width=0.75\linewidth]{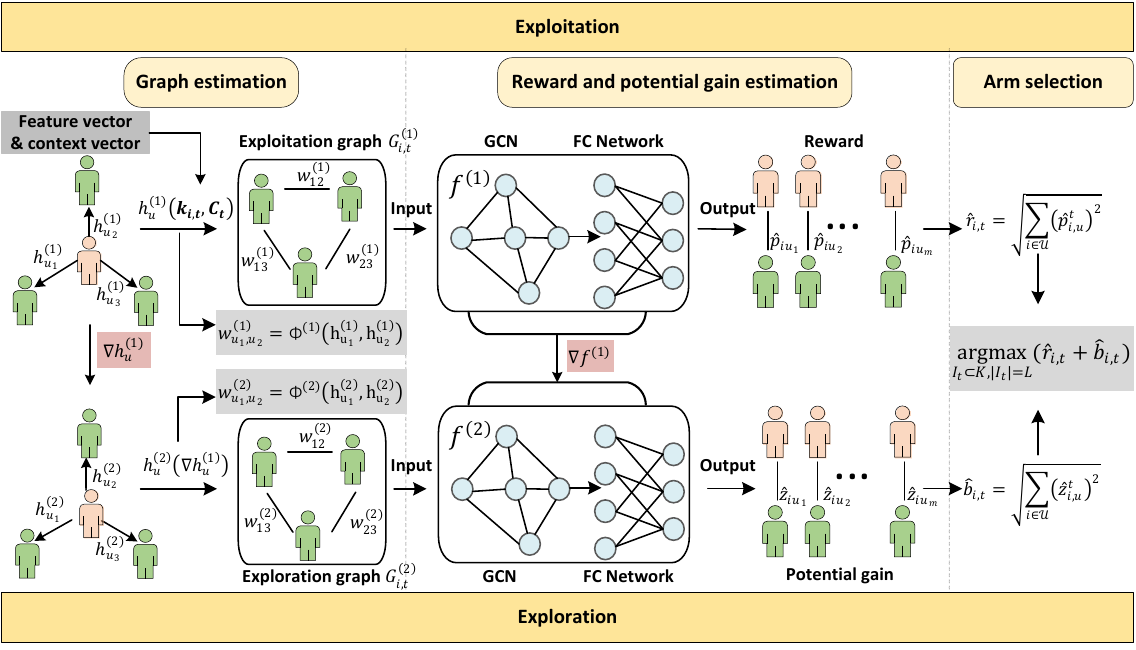}
   \vspace{-2mm}
  \caption{The framework of IM-GNB. For each arm, we initially take the arm feature vector and the current context vector $(\bm{k}_i,\bm{C}_t)$ as inputs to estimate the diffusion probability for each user-arm pair with $h_u^{(1)}$. Subsequently, we assess the potential gain on the diffusion probability with the past gradient of $h_u^{(1)}$, yielding both exploitation and exploration graphs. 
  With the pre-estimated graphs, we refine the estimate of the diffusion probability for each user-arm pair with $f^{(1)}$ and $f^{(2)}$. The aggregate reward of the arm across all users is derived from the sum of all the refined individual diffusion probabilities. The potential gain is measured similarly. Finally, we select the arm with the highest sum of estimated reward and its potential gain.}
  \label{fig:framework}
\end{figure*}

Many recent works~~\cite{iacob2022contextual,wen2017online} on the IM problem that exploit bandits for the exploration-exploitation trade-off assume that the reward is a linear or generalized linear function of arm vectors. Considering the high complexity and dynamicity of social network-related data, we use the representation power of neural networks to firstly, learn users' connectivity to build exploitation and exploration graphs and secondly, learn the underlying reward function and the potential gains on the estimated reward. The overall framework of our model is illustrated in Fig.~\ref{fig:framework}. 

\subsection{Estimating the User Graphs}
\label{sec:pre-estimation}
In this section, we first provide a strategy to estimate the users' correlations to be influenced by the same arm, forming the basis for the exploration-exploitation strategy in Sec.~\ref{sec:GNB}.
\vspace{-1mm}
\subsubsection{User exploitation graph}
We bridge the users in the social graph with diffusion probabilities between influencers and users. The intuition is that given the same message to be diffused (context $C_t$), users who exhibit high correlations in this graph are more likely to be influenced by the same influencer. As the context changes, an influencer may not exert the same influence on users. 
Thus, at each round $t$ and for each influencer (arm) $k_i$, we induce an exploitation graph $G_{i,t}^{(1)}$ to represent the users' correlations.

In the exploitation graph $G_{i,t}^{(1)}$, the weights among users are referred to as {\em users' correlations} w.r.t. the diffusion probability from \ arm $k_i$ (hence likelihood to influenced by the same influencer $k_i$). For each user $u\in \mathcal U$, we use a neural network as the pre-defined hypothesis function $h_u^{(1)}=h_u^{(1)}(\bm{k}_i, \bm{C}_t;[\mathcal{P}_u^{(1)}]_{t-1})$ to learn these probabilities ($[\mathcal{P}_u^{(1)}]_{t-1}$ are the updated parameters of the network from round $t-1$). The weights in $G_{i,t}^{(1)}$ are 
\begin{equation}
        w_{i,t}^{(1)}(u,u') = \Phi^{(1)} \big(
        h_u^{(1)}(\bm{k}_i, \bm{C}_t), 
        h_{u'}^{(1)}(\bm{k}_i, \bm{C}_t)\big), 
    \end{equation}
where $\Phi^{(1)}$ is the same function in Eq.~\eqref{eq: weights}. For each user $u$, $h_u^{(1)}$ will be trained by gradient descent (GD) with the given context and the chosen arm as input and the reward as label. The loss is defined as
\begin{equation}
\label{eq:distinctive}
    \mathcal{L}_u^{(1)} = \Big(h_u^{(1)}\big(\bm{k}_i, \bm{C}_t; \mathcal{P}_u^{(1)}\big)-d_u^t\Big)^2,
\end{equation}
where $d_{u}^t =1$ if $c_u^t>0$ and $c_u^{t-1}=0$, else $d_{u}^t =0$.
Recall that we use $c_u^t$ to denote the total number of times user $u$ has been activated (influenced) up to and including round $t$, and we only count the newly activated nodes at each round.

\vspace{-1mm}
\subsubsection{User exploration graph}
\label{sec:pre-estimation_explore}
Recent works on  neural bandits~\cite{ban2021ee,zhou2020neural,zhang2020neural,ban2021multi,ban2021local} take advantage of the representation power of  neural networks to learn the uncertainty of estimation (potential gain). 
These works use the past gradient to incorporate the feature of arms and the learned discriminative information of estimation function ($h^{(1)}_u(\bm{k}_i,\bm{C}_t)$ in our work). 


Qi et al.~\cite{qi2023graph}   applied this paradigm in collaborative filtering for user-item pair prediction in online recommendation scenarios and demonstrated its effectiveness. Since the IM problem shares similarity with predicting user preferences towards items (in our case susceptibility to  influencers), especially when the connections (correlations) among users are reinforced by social ties, we  apply   the past gradient to quantify the ``exploration bonus''~\cite{lattimore2020bandit}.



For a   user $u\in\mathcal{U}$, we use a neural network   $h_u^{(2)}$ to learn the uncertainty of the estimated diffusion probability between arm $k_i$ and user $u$, i.e., $\mathbb{E}[p_{i,t}|u,\bm{k}_i,\bm{C}_t]-h_u^{(1)}(\bm{k}_i,\bm{C}_t)$, similar to Eq.~\eqref{eq:weight2}. As in~\cite{ban2021ee}, we apply $h_u^{(2)}$ directly on the previous gradient of $h_u^{(1)}$. Analogously, the exploration graph $G^{(2)}_{i,t} = ({\mathcal{U}}, E, W^{(2)}_{i,t})$ is constructed with $W^{(2)}_{i,t}=\big\{w_{i,t}^{(2)}(u,u'):u,u'\in\mathcal{U}\big\}$, and $w_{i,t}^{(2)}(u,u')$ is the exploration correlation among users, defined as
\begin{equation}
    w_{i,t}^{(2)}(u,u') = \Phi^{(2)}\Big(h^{(2)}_u\big(\nabla h_u^{(1)}\big),h^{(2)}_{u'}\big(\nabla h_{u'}^{(1)}\big)\Big).
    \vspace{-1mm}
\end{equation}
The previous gradient $\nabla h_u^{(1)}(\bm{k}_i,\bm{C}_t)=\nabla_{\mathcal{P}}h_u^{(1)}(\bm{k}_i,\bm{C}_t;{[\mathcal{P}_u^{(1)}]}_{t-1})$ is the network gradient at round $t -1$, with ${[\mathcal{P}_u^{(1)}]}_{t-1}$ the last updated parameters of $h_u^{(1)}$. In addition, $\Phi^{(2)}$ is the function defined in Eq.~\eqref{eq:weight2} and $h_u^{(2)}$ will be trained with GD, where the previous gradient of $h_u^{(1)}$ is computed based on the input samples, and the residual diffusion probability (potential gain on the estimated diffusion probability) is the label, with the loss given as
\begin{equation}
    \mathcal{L}_u^{(2)} = \Big(h^{(2)}_u\big(\nabla h_u^{(1)}(\bm{k}_i,\bm{C}_t)\big) - \big(d_u^t-h_u^{(1)}(\bm{k}_i, \bm{C}_t)\big)\Big)^2.
    \vspace{-1mm}
\end{equation}


Regarding the network structure of $h^{(1)}$ and $h^{(2)}$, since there are no data characteristics requiring specific models such as RNNs for sequential dependencies or CNNs for visual content, we simply employ an
$\LL$-layer fully connected (FC) neural network at this stage for initial graph estimation.

To summarise, we use $h_u^{(1)}$,
denoting user $u$, to obtain the estimated diffusion probability from influencer (arm) $k_i$ to $u$ (the estimation function is built for each user individually, i.e., there are $m$ estimation functions $h_u^{(1)}$ in total), and the \textbf{exploitation graph} $G_{i,t}^{(1)}$ for arm $k_i$ is built such that the basic users are correlated based on estimated diffusion probabilities. We also apply $h_u^{(2)}$ 
to represent the uncertainty of the estimated diffusion probability, 
and the \textbf{exploration graph} $G_{i,t}^{(2)}$ corresponding to arm $k_i$ is built such that the users are correlated based on the potential gains. The graph estimation process is given in Lines~\ref{algo:graph_construction}--\ref{algo:final_return} of  Algorithm~\ref{algo:im_gnb}.

\subsection{Exploitation-Exploration with GNNs}
\label{sec:GNB}
With the   user correlation graphs $G_{i,t}^{(1)}$ and $G_{i,t}^{(2)}$  for \emph{exploitation} and   \emph{exploration} respectively, we now have a refined estimate of the diffusion probabilities between influencers and users, as well as the expected total spread (newly activated users), i.e., the reward.
\subsubsection{GNN for exploitation}
In round $t$, for each arm $k_i$, with the pre-estimated exploitation graph $G_{i,t}^{(1)}$ for arm $k_i$ as input, we use a GNN model $ f^{(1)}(\bm{k}_{i},\bm{C}_t,G_{i,t}^{(1)};\mathcal{P}^{(1)})$ to estimate the reward described in Eq.~\eqref{eq:reward_total}, with $\mathcal{P}^{(1)}$ representing the parameters of $f^{(1)}$. 

We define for each arm a symmetric adjacency matrix $A_{i,t}^{(1)}\in\mathbb{R}^{m\times m}$ from the exploitation graph $G_{i,t}^{(1)}$, with each element 
in the matrix corresponding to the correlations weights $w_{u,u'}$ between user $u$ and user $u'$ in $G_{i,t}^{(1)}$, and the normalized adjacency matrix~\cite{kipf2016semi} being $S_{i,t}^{(1)} = D^{-\frac{1}{2}}A_{i,t}^{(1)}D^{-\frac{1}{2}}$, with $D$  the degree matrix. We concatenate the arm feature vector $\bm{k}_i$ with the context vector $\bm{C}_t$ to build the feature matrix $\bm{X}_{i,t} = \text{diag} ([\bm{k}_{i},\bm{C}_t], [\bm{k}_{i},\bm{C}_t], \ldots, [\bm{k}_{i},\bm{C}_t]) \in \mathbb{R}^{m\times m(d_1+d_2)}$.


We adopt a simplified Graph Convolutional Network   (GCN) model~\cite{wu2019simplifying} to learn the aggregated representation of the exploitation graph. With $S_{i,t}^{(1)}$ and $\bm{X}_{i,t}^{(1)}$ as inputs, the feature representation matrix is expressed as
\begin{equation}
\label{eq:GCN}
    H_{\mathcal{G}} = \sigma\Big(\big(S_{i,t}^{(1)}\big)^\gamma \bm{X}_{i,t}^{(1)};\mathcal{P}^{(1)}_{\mathcal{G}}\Big),
    \vspace{-1mm}
\end{equation}
where $\sigma$ is the activation function, $\mathcal{P}^{(1)}_{\mathcal{G}}\in \mathbb{R}^{m(d_1+d_2)\times p}$ is the trainable weight matrix in the GCN model, and $\gamma$ is the number of hops the information propagating over the user graph, indicating that  after $k$ layers a node obtains the feature information from all nodes found $\gamma$ hops away in the graph. In the GCN model, $\bm{X}_{i,t}^{(1)}$ is applied to the corresponding weight matrix $\mathcal{P}^{(1)}_{\mathcal{G}}$ so that $\mathcal{P}^{(1)}_{\mathcal{G}}$ is partitioned for each user $u\in \mathcal{U}$ to get the $p$-dimensional arm-user diffusion representation, corresponding to each row of $H_{\mathcal{G}}\in \mathbb{R}^{m\times p}$.

To further refine the $p$-dimensional arm-user pair representation in $H_{\mathcal{G}}$, we add an $\LL$-layer FC neural network to the GCN model, and for $l\in \{1, 2, \ldots, \LL-1\}$ the representation for each layer is
\begin{equation}
\label{eq:FC}
    H_l = \sigma\big(H_{l-1} \cdot \mathcal{P}_{l}^{(1)}\big),\quad \mbox{and}\quad H_0=H_{\mathcal{G}},
\end{equation}
with $H_l\in \mathcal{R}^{m\times p}$, and $\mathcal{P}_l^{(1)}$ being the trainable parameters in each layer in the FC network. For the last layer, we have
\vspace{-2mm}
\begin{equation}
\vspace{-1mm}
\label{eq:reward_estimation}
    \hat{P}_{i,t}=H_{\LL-1}\cdot\mathcal{P}_{\LL}^{(1)},
    \vspace{-1mm}
\end{equation}
where the $\mathcal{P}_{\LL}^{(1)}$ are the parameters in the last layer, and $\hat{P}_{i,t}\in \mathbb{R}^m$ is the $m$-dimensional vector with each element the refined estimated diffusion probability $\hat{p}_{i,u}^t\in \mathbb{R}$ between arm $k_i$ and   user $u \in \mathcal{U}$.

With the refined estimated diffusion probability between arm $k_i$ and all the users, the estimated reward for arm $k_i$ across all the users is computed as the norm of the output layer:
\begin{equation}
\label{eq:norm_1}
    \hat{r}_{i,t} = \| \hat{P}_{i,t} \|=\sqrt{\sum_{u\in \mathcal{U}}\big(\hat{p}_{i,u}^t\big)^2}.
\end{equation}
The exploitation network will be trained with GD based on the influence spread from arm $k_i$, where the predicted output are the diffusion probabilities across all users, with the label (reward)
\begin{equation}
\label{eq:reward2}
    r_{i,t}=\sum_{u\in{A_{k_i}}}d_u^t.
\end{equation}
Recall that $A_{k_i}$ is the set of users activated or influenced by arm $k_i$ at round $t$, and $d_u^t$ is defined as the distinct activations  in Eq.~\eqref{eq:distinctive}. 
For a refined learning on each user-arm diffusion pair, we calculate  the quadratic loss w.r.t.\ each user individually, as
\begin{equation}
    \mathcal{L}^{(1)} = \sum_{u\in{A_{k_i}}}\big(\hat{p}_{i,u}^t -d_u^t\big)^2.
\end{equation}

\subsubsection{GNN for exploration}
\label{sec:GNN_explore}
Similar to the user graph pre-estimation described in Sec. ~\ref{sec:pre-estimation}, we follow the exploration-exploitation strategy by applying a gradient-based exploration function w.r.t.\ the exploitation function; also see~\cite{ban2021ee,qi2023graph,zhou2020neural,zhang2020neural} for similar strategies.

In round $t$, for each arm $k_i$, with the induced exploration graph $G_{i,t}^{(2)}$  where the pre-estimated weights in Sec.~\ref{sec:pre-estimation_explore} represent the exploration correlations among users, we apply another GNN model $ f^{(2)}(\nabla f^{(1)},G_{i,t}^{(2)} ;\mathcal{P}^{(2)})$ to evaluate the potential gain (the gap between expected reward and estimated reward) for arm $k_i$, where $\nabla f^{(1)}=\nabla_{\mathcal{P}}f^{(1)}(\bm{k}_i,\bm{C}_t;{[\mathcal{P}^{(1)}]}_{t-1})$, and $\mathcal{P}^{(1)}$ and $\mathcal{P}^{(2)}$ are the parameters of $f^{(1)}$ and $f^{(2)}$ respectively. 

We adopt the same network architecture as in the exploitation network to learn the representation matrix for the exploration graph with a $k$-hop simplified GCN, and to predict the potential gain with an $\LL$-layer FC neural network. The architecture of $f^{(2)}$ can be also implemented via Eqs.~\eqref{eq:GCN}--\eqref{eq:reward_estimation}, with the input feature vector $\bm{X}_{i,t}^{(2)}\in \mathbb{R}^{m\times mq}$ and trainable matrix $\mathcal{P}^{(2)}_{\mathcal{G}}\in \mathbb{R}^{mq\times p}$ in the GCN model, where $q$ is the dimension of input gradient. In the GCN of $f^{(2)}$, the input gradient matrix $\bm{X}_{i,t}^{(2)}$ is similarly applied to partition the weight matrix $\mathcal{P}^{(2)}_{\mathcal{G}}$, so that  each user-arm pair is represented by a $p$-dimensional vector   for  the purpose of exploration.

In the output layer we obtain an $m$-dimensional vector $\hat{Z}_{i,t}$, where each element represents the estimated potential gain $\hat{z}_{i,u}^t\in \mathbb{R}, u\in \mathcal{U}$ (with $|\mathcal{U}|=m$) for each user-arm pair. 
With the estimated potential gains from the output layer, the overall estimated potential gain for arm $k_i$ is obtained as the norm of output $\hat{Z}_{i,t}$, i.e., 
\begin{equation}
\label{eq:norm_2}
    \hat{b}_{i,t} = \| \hat{Z}_{i,t} \|=\sqrt{\sum_{u\in \mathcal{U}}\big(\hat{z}_{i,u}^t\big)^2}.
\end{equation}

When training $f^{(2)}$ with GD, the quadratic loss is computed between the estimated potential gain and the residual gain (the gap between the reward in Eq.~\eqref{eq:reward2} and the estimated reward), as
\begin{equation}
    \mathcal{L}^{(2)} = \sum_{u\in{A_{k_i}}}\Big(\hat{z}_{i,u}^t - \big(d_u^t-\hat{p}_{i,u}^t\big)\Big)^2.
\end{equation}

The computations of the reward and potential gain are summarized in Lines~\ref{algo:reward}--\ref{algo:gain} in Algorithm~\ref{algo:im_gnb}.
\subsubsection{{\modelname} arm selection}
\begin{algorithm}
\SetKwFunction{proc}{Estimating graphs for arm}
  \SetKwProg{myalg}{Algorithm}{}{}
  \SetKwProg{myproc}{Procedure}{}{}
		\caption{IM-GNB}
        \label{algo:im_gnb}
		\KwInput{ Influencer set $K$, number of selections $L$ per round} 
        \KwOutput{Arm recommendation for each time step $t$}
        Initialization of all the trainable parameters\\
		\For{$t = 1,2,3, \dots,T$}
		{
			Receive from environment the context $\bm{C}_t$\\
                \For{$k_i\in K$}
                {
                    \state construct two user graphs $G_{i,t}^{(1)}$ and $G_{i,t}^{(2)}$ from 
                    \textbf{Procedure} \textit{Estimating
                    graphs for arm $k_{i}$}\\
                    \state \label{algo:reward} Compute estimate of reward 
                    $\hat{r}_{i,t} = f^{(1)}(\bm{k}_{i},\bm{C}_t,G^{(1)}_{i,t};[\mathcal{P}^{(1)}]_{t-1})$ \\
                    and \state \label{algo:gain} potential gain
                    $\hat{b}_{i,t} = f^{(2)}(\nabla [f^{(1)}]_{i,t},G^{(2)}_{i,t};[\mathcal{P}^{(2)}]_{t-1})$

                }
			\state \label{algo:arm_select} choose arm set $I_t=\arg\max_{I_t\subset K,|I_t|=L}(\hat{r}_{i,t}+\hat{b}_{i,t})$ and observe the true reward (spread) $R_t$ in Eq.~\eqref{eq:true_spread}, which represents the newly activated users.\\
            \For{$u \in \mathcal{U}$}
            {
            train the user networks $h^{(1)}_{u}(\cdot ; \mathcal{P}^{(1)}_{u})$, $h^{(2)}_{u}(\cdot ; \mathcal{P}^{(2)}_{u})$ }
            \For{$k \in \mathcal{K}$}
            {
            train the GNN models $f^{(1)}(\cdot;\mathcal{P}^{(1)})$,$f^{(2)}(\cdot;\mathcal{P}^{(2)})$ }

		}
    
    \myproc
    {\state \label{algo:graph_construction}  \proc{$k_{i,t}$}}
  {
  \For{each user pair $(u,u')\in \mathcal U \times \mathcal U$}
  {
  \textbf{For} edge weight $w^{(1)}_{i,t}(u,u')\in W^{(1)}_{i,t}$, \textbf{update}\
  $w^{(1)}_{i,t}(u,u')=\Phi ^{(1)}(h^{(1)}_u(k_{i,t}),h^{(1)}_{u'}(k_{i,t}))$\\
  \textbf{For} edge weight $w^{(1)}_{i,t}(u,u')\in W^{(1)}_{i,t}$, \textbf{update}\
  $w^{(2)}_{i,t}(u,u')=\Phi ^{(2)}(h^{(2)}_u(\nabla h^{(1)}_u),h^{(2)}_{u'}(\nabla h_{u'}^{(1)}))$\\
  \KwRet  \label{algo:final_return}   $G_{i,t}^{(1)}$ and $G_{i,t}^{(2)}$
  }
  
  }
\vspace{-1mm}
\end{algorithm}

We summarize the {\modelname} framework in Algorithm~\ref{algo:im_gnb}. For an information diffusion campaign with $T$ rounds, we select $L$ influencers (arms) from a known influencers base $K$ at each round $t$ to diffuse the given message $\bm{C}_t$. At each round $t$ for each arm $k_i$, we firstly construct the two user graphs i.e., the \emph{exploitation graph} and the \emph{exploration graph} via a procedure (Lines~\ref{algo:graph_construction}--\ref{algo:final_return}) of pre-estimation on graph weights, which capture users' correlations in terms of exploitation and exploration respectively. With the derived graphs, we compute the overall expected reward $\hat{r}_{i,t}$ and potential gain $\hat{b}_{i,t}$ for each arm in Eqs.~\eqref{eq:norm_1} and~\eqref{eq:norm_2}, as the norms of the output vectors from $f^{(1)}$ and $f^{(2)}$. 
Next, we select the arm set based on the maximum of the sum of reward estimation and potential gain $\hat{r}_{i,t}+\hat{b}_{i,t}$ (Line 8). Finally, for each user $u\in \mathcal{U}$, we train the user's neural networks from pre-estimation, and we train for each arm $k_i\in K$ the GNN models (Lines 9--12).

We observe from the above that at each round $t$, $h_u^{(1)}$ will take as input the feature vector of a certain arm $k_i$, along with the context vector $\bm{C}_t$ to provide an initial estimate on the diffusion probability for user $u$ being influenced by arm $k_i$. Subsequently, the gradient of $h_u^{(1)}$ is employed as input to estimate the potential gain in diffusion probability. Parameters in $h_u^{(1)}$ and $h_u^{(2)}$ undergo continuous training and updating at each round to refine the approximation functions for user $u$, predicting the probability of being influenced by any arm within various contexts. Similarly, for the reward estimation for each arm with $f^{(1)}$ and $f^{(2)}$, $f^{(1)}$ takes as input the arm feature vector and context vector, as well as the pre-estimated graph $G_{i,t}^{(1)}$ to refine the initial estimation on the diffusion probability, allowing to estimate the reward across all users, and the gradient of $f^{(1)}$ serves as the input of exploration function $f^{(2)}$. Both $f^{(1)}$ and $f^{(2)}$ undergo continuous training to refine the reward estimation function (exploitation) and potential gain (exploration). This iterative process ensures that $f^{(1)}$ and $f^{(2)}$ adapt effectively to diverse contexts and user correlation graphs.

\subsubsection{Complexity Analysis}
Recall from the previous notation conventions that we have $|K|=n$   arms, $m$  users, and the dimensions of the feature vectors and context information are $d_1$  and $d_2$ respectively. For simplicity, we use $d_g$ to denote the dimension of all the input gradients, and we assume that  the same  structure is used for all the FC neural networks in our model. In particular, each neural network has $\LL$ layers and each layer has $n$ neurons.

For the pre-estimation of user exploitation and exploration graphs, at each round, the complexity of the pre-defined hypothesis functions $h_u^{(1)}$ and $h_u^{(2)}$ (FC neural networks) is $O\big(|K|m\LL (d_1+d_2)n\big)$ for exploitation and $O(|K|m\LL d_gn)$ for exploration.

For the refined estimation procedure where we use GCNs, as we predict correlations among all users, the graphs can be seen as complete. Assuming that the exploration / exploitation GCNs share the same NN structure, the time and space complexities for exploitation are $O\big(|K|\LL  (m^2(d_1+d_2)+m(d_1+d_2)^2 )\big)$ and $O\big(|K|\LL  (m^2+(d_1+d_2)^2+m(d_1+d_2) )\big)$ respectively, and for exploration $O\big(|K|\LL (m^2d_{g}+md_{g}^2)\big)$ and $O\big(|K|\LL (m^2+d_{g}^2+md_{g})\big)$.


From  the above discussion, we can observe that the number of users and the dimension of the input gradient are the most critical parameters in determining the time complexity. We consider the following methods to reduce the computational complexity. 

\paragraph{User clustering}

In applications with billions of users on social media, it is impractical and  excessively costly to predict diffusion probabilities  and correlations   at the granularity of individual users. In response to this challenge, we can leverage the posting activity (e.g., retweeting history) of users to construct a topic distribution vector for each user. We can then  cluster users into a specific number of groups, with each user group representing a macro-node in a smaller social graph. Notably, the theoretical underpinnings outlined earlier remain applicable to these clustered user groups, and user $u$ becomes $u_{c_i}, i =1,2,\ldots, m'$ 
and $\cup_{i=1}^{m'}u_{c_i}=\mathcal{U}$, with $m'$ denoting the number of clustered groups. Despite this adjustment, we continue to refer to the user group $u_{c_i}$ as user $u$ for simplicity. 

The introduction of clustering can significantly reduce computational cost, transforming the space complexity of the adjacency matrix in the GCN  from $m\times m$ to a more computationally efficient scale. Experiments are carried out on the number of clustering groups in Sec.~\ref{sec:hyper_cluster} to show the impact of the number of clusters on the performance of the model.

\paragraph{Input gradients}
In Sec.~\ref{sec:pre-estimation_explore} and  Sec.~\ref{sec:GNN_explore}, we saw that the input dimensions for the previous gradients can pose computational challenges due to their potentially large values. This is particularly relevant in Sec.~\ref{sec:GNN_explore}, where the input gradient dimension is $m(d_1+d_2) p+(\LL-1)p^2 + p$. To address this issue, and inspired by approaches commonly employed in CNN-related works, we use the {\em  average pooling} technique to effectively reduce the input dimension and improve efficiency.

 \vspace{-1mm}
\section{Experiments}
 \vspace{-1mm}
\label{sec:experiments}

 In this section, we evaluate  our model {\modelname} on  datasets  from Twitter and Sina Weibo. We compare it with baselines also designed for multi-round diffusion campaigns, and we analyze the comparison results in the end. For reproducibility, the {\modelname} code is available at {\bf \url{https://github.com/goldenretriever-5423/IM_GNB}}.

\noindent \textbf {Datasets}
Twitter and Weibo are two of the largest social media platforms. We collected the Twitter dataset through its API. In our context analysis for Twitter, we apply  $K$-means clustering on the public vocabulary \emph{glove-twitter-200}~\cite{pennington2014glove}, available from the Gensim word embedding open-source library\footnote{https://pypi.org/project/gensim/}. The resulting clusters provide centroids that serve as representative themes within the dataset. Subsequently, we represent them as a distribution across these centroids ($10$ in our experiments) to encode tweets. Each word in a tweet is assigned to its nearest centroid, resulting in the overall distribution. The feature vector of the influencer is the normalized aggregation of all its historical tweets. The Weibo dataset~\cite{weibopaper} is a publicly available one built for information diffusion studies. In this dataset, each  post is encoded with a distribution over the 100 topics~\cite{weibopaper} using latent Dirichlet allocation~\cite{blei2003latent}. Similar to the Twitter dataset, the feature vector of the influencer is the normalized aggregate of the topic distribution of all  historical tweets. 

To simulate   campaigns on social media, we assume that the marketer has access to only a few most important influencers to diffuse the message in the campaigns. Hence, we fix the size of the influencer set $K$ by selecting the users with the highest number of reposts in our  Twitter and Weibo logs, and we keep all the tweets related to them. The statistics of the datasets before and after filtering are given in Tables~\ref{tab:tab_datasets_orig} and~\ref{tab:tab_datasets_filtered}. In each campaign, we randomly chose the contexts (tweets), referred to as {\em  topic distributions}, for each round from the  pool of available contexts within the dataset.
 \vspace{-1mm}
\begin{table}[!htb]
  \caption{Description of original datasets.}
  \vspace{-2mm}
  \label{tab:tab_datasets_orig}
  \begin{tabular}{lccc}
    \toprule
         & {\#users} &{ \#original tweets}& \# retweets\\
    \midrule
          {\itshape Twitter} & 11.6M& 242M& 341.8M \\
          {\itshape Weibo} & 1.8M & 300K & 23.8M\\
  \bottomrule
\end{tabular}
\end{table}


\begin{table}[!htb]
\vspace{-2mm}
  \caption{Description of  filtered datasets.}
  \vspace{-2mm}
  \label{tab:tab_datasets_filtered}
  \begin{tabular}{lccc}
    \toprule
         & {\#users} &{ \#original tweets}& \# retweets\\
    \midrule
          {\itshape Twitter} & 31.6k& 19k& 1M \\
          {\itshape Weibo} & 54.4k & 6K & 1M\\
  \bottomrule
\end{tabular}
\end{table}

\noindent \textbf{Baselines}
We compare {\modelname} to a set of bandit models designed for the IM problem under the same multi-round campaigns scenario, where the underlying network is unknown and no assumptions about the diffusion models are made. LogNorm-LinUCB~\cite{iacob2022contextual} and GLM-GT-UCB~\cite{iacob2022contextual} are the recent, state-of-the-art approaches to maximize information diffusion during such IM campaigns. LogNorm-LinUCB directly adapts the LinUCB algorithm by using logarithmic normalization and contextual information to make sequential selections of spread seeds, while GLM-GT-UCB employs a generalized linear model and the Good--Turing estimator to determine the remaining potential of influencers. We also compare with FAT-GT-UCB~\cite{lagree2018algorithms}, a context-free model which has the particularity to consider the \emph{fatigue}, i.e., an influencers' diminishing tendency to activate basic users as they are re-seeded throughout the campaign. We also generalize several  state-of-the-art neural bandit methods--NeuralUCB~\cite{zhang2020neural} and NeuralTS~\cite{zhou2020neural},  as well as the well-known LinUCB~\cite{chu2011contextual}   to our multi-round IM campaign. Finally, we also implement a reference model that randomly chooses the influencer(s) at each round, as in~\cite{iacob2022contextual}.

\noindent \textbf{Experimental Setting}
In our experiments, to reduce the computation cost,  we first cluster all the users into $50$ groups. For the pre-estimation of the graph weights, we use a 3-layer FC neural network as the hypothesis class for both $h_u^{(1)}$ and $h_u^{(2)}$, to estimate the diffusion probability and potential gain. The functions $\Phi^{(1)}$ and $\Phi^{(2)}$ that map the users' correlations to the weights in the graphs are radial basis functions (RBFs), with their bandwidths set to $5$. For the GCN model, we explore the use of 3 hops (i.e., $\gamma = 3$) to capture multi-level relationships within the user graphs, with a 3-layer FC neural network connected at the end. The pooling step sizes to reduce dimensions~\cite{qi2023graph} for the input gradients in $f^{(2)}$ are set to $1,000$ and $10,000$ respectively for   the Twitter and Weibo datasets. 

\noindent \textbf{Empirical Results}
For a diffusion campaign, at each round, the environment first provides the context,  an algorithm then selects the round's  influencer(s), and finally, a tweet is sampled for the specific pair of influencer(s) and context from the dataset. The new activations are determined by discounting the users previously encountered from the set of users associated with the sampled tweet. All our empirical  results are averaged over 100  independent runs (the means and standard deviations are reported), and the diffusion budget is set to 500 rounds.
\begin{figure*}[t!]
  \centering
  \includegraphics[width=0.9\linewidth]{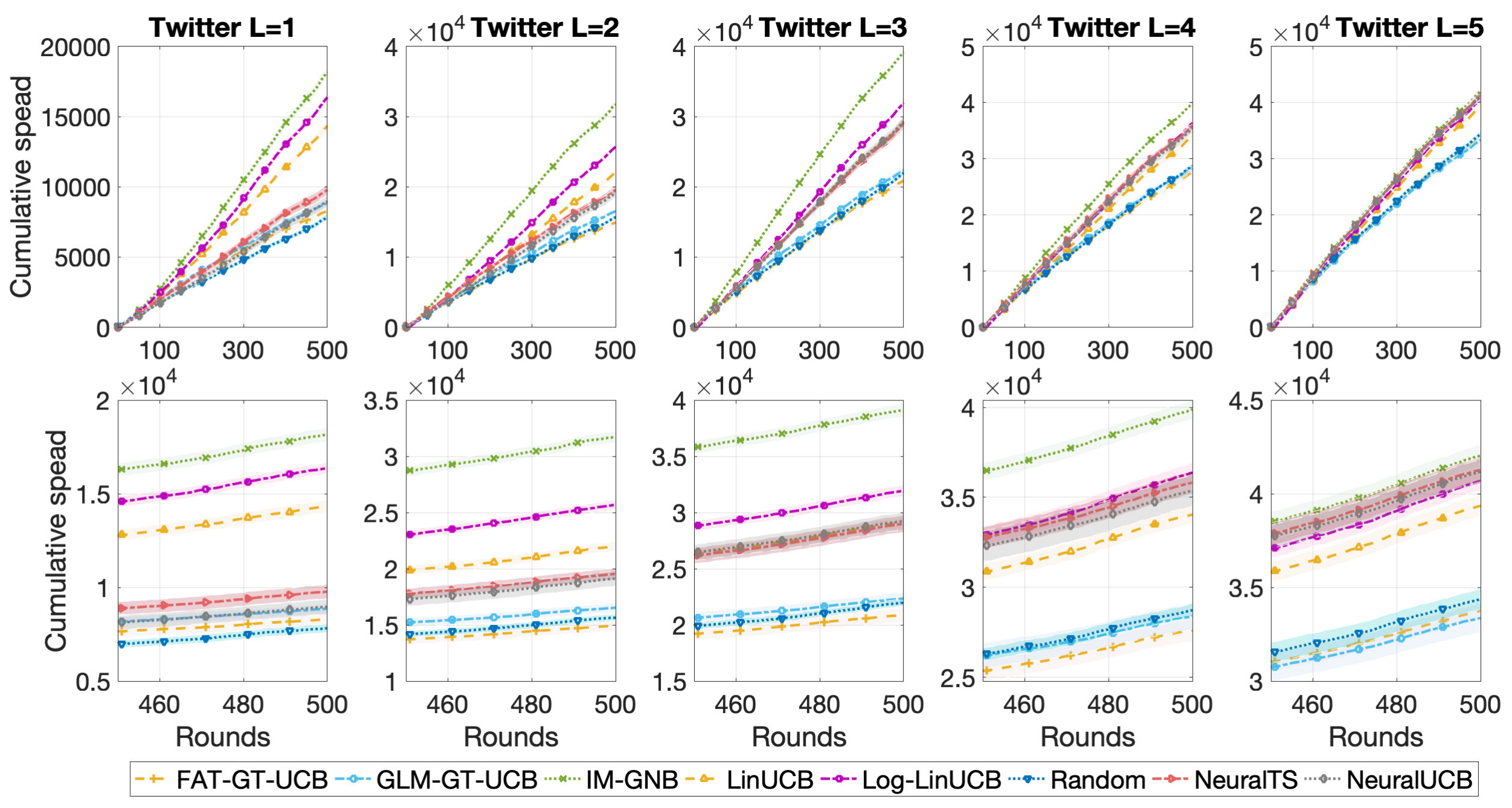}
  \vspace{-2mm}
  \caption{Comparison of IM-GNB with  baselines on the Twitter dataset.}
  \vspace{-3mm}
\label{fig: baseline_twitter}
\vspace{-1mm}
\end{figure*}

\begin{figure*}[t]
  \centering
  \includegraphics[width=0.9\linewidth]{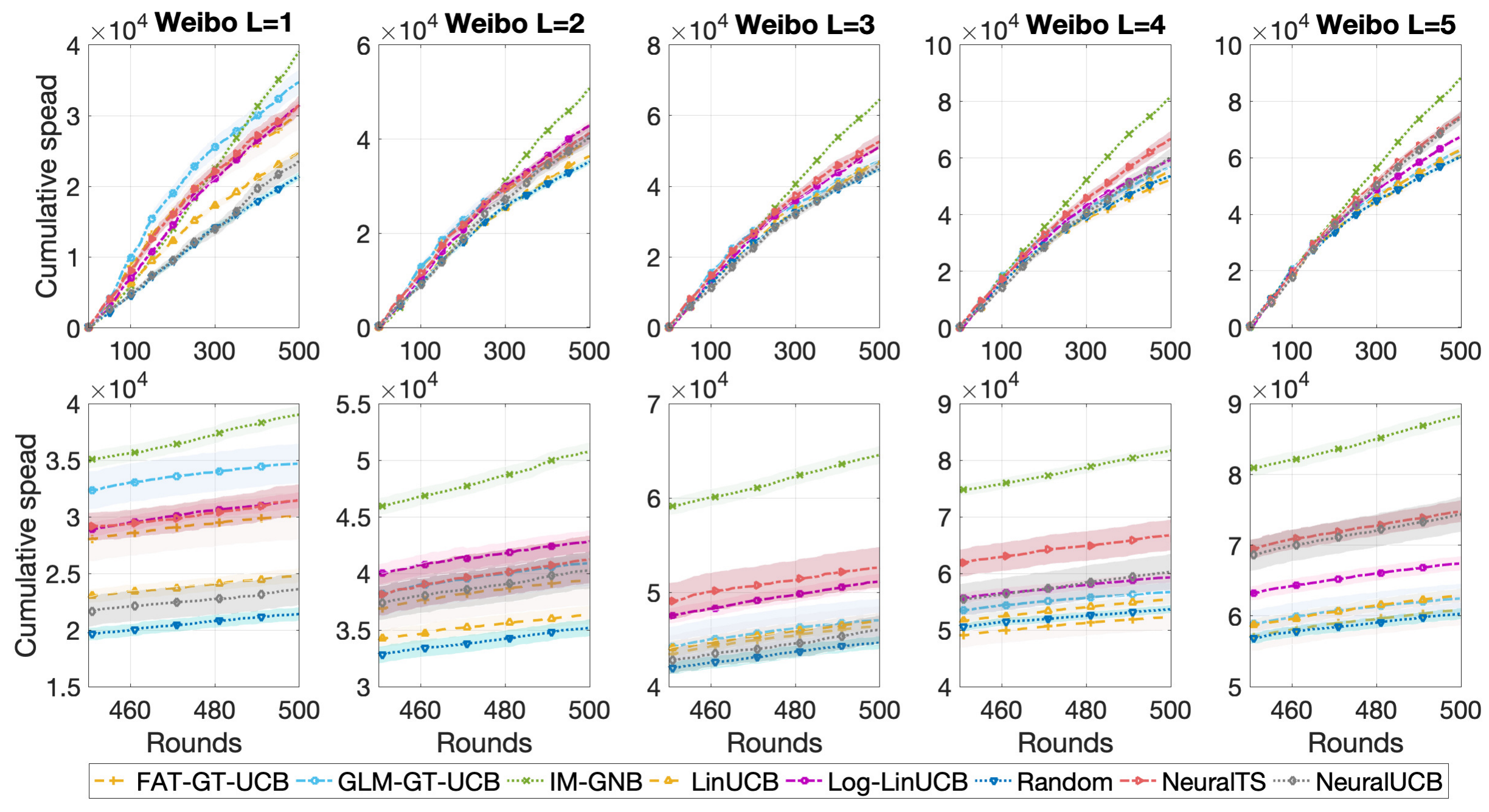}
  \caption{Comparison of IM-GNB with baselines on the Weibo dataset.}
\label{fig: baseline_weibo}\vspace{-3mm}
\end{figure*}

\noindent \emph{Comparison with baselines:}
We conducted comparisons with various baselines on the Twitter and Weibo datasets, varying the number of chosen seeds ($L$) per round within  $\{1,2,\ldots,5\}$. The results are shown in Fig.~\ref{fig: baseline_twitter} and Fig.~\ref{fig: baseline_weibo} respectively. From the two figures, we can observe that, across both datasets, our model IM-GNB generally outperforms the baselines. Notably, on the Twitter dataset, IM-GNB exhibits a significantly increased advantage over the baselines, as the number of seeds increases up to $L=3$. However, this advantage diminishes as $L$ continues to grow, as the probability of selecting the correct arms increases for all models. In fact, for the extreme scenario where $L=|K|=10$, this results  in the same performances across all models due to the selection of the entire base set of seeds. Similarly, for the Weibo dataset, IM-GNB demonstrates its largest advantage at $L=4$. These results validate our motivation to leverage the expressivity of both neural networks and bandit algorithms in IM campaigns, enabling us to effectively capture dynamic user-user and user-influencer interactions using  GNBs.

It is worth noting that, in the Weibo dataset, particularly when $L$ is small (e.g., $L=1,2$), the performance in the initial rounds is surpassed by certain baseline methods, notably GLM-GT-UCB, which exhibit more rapid learning capabilities. We attribute this phenomenon to the nature of IM-GNB as a data-driven model, typical of modern deep learning-based approaches. The efficiency of IM-GNB improves rapidly with the accumulation of data (i.e., as more rounds pass by), indicating potential slower convergence initially, but yielding better results as the number of rounds increases. Additionally, the Weibo dataset is a publicly available dataset that consists of artificially extracted data from diffusion cascades, while the Twitter dataset, albeit sparser than Weibo, offers insights closer to real-world IM  scenarios.

In both datasets,   Lognorm-LinUCB generally outperforms the other baselines. While there are instances where GLM-GT-UCB briefly outperforms Lognorm-LinUCB in the initial stages, the latter demonstrates stable performance with smaller error bars. This underscores the robustness of the log-normal  assumption on the reward distribution. NeuralUCB and NeuralTS, as scalarizations of the general neural bandit model, exhibit comparable performances across both datasets. Notably, their effectiveness lags behind models tailored for multi-round diffusion campaigns when $L$ is small. However, when $L$ increases, the models are empowered with more data, showing  marked performance improvements.

\noindent \textbf{Hyperparameter Analysis}

\noindent \emph{{Number of clusters:}}
\label{sec:hyper_cluster}
We conduct experiments on the number of clusters in the Twitter dataset with $L=2$, and the results on the last round (final accumulated spread) are shown in Fig.~\ref{fig:group} of the appendix. We observe from Fig.~\ref{fig:group}  that the campaign performance improves as the number of clusters increases from $2$ to $150$ at the beginning. However, beyond $200$ clusters, performance begins to decline. This decline can be likely attributed to insufficient data within each cluster for effective learning with the constraints of a limited budget on the number of rounds. Additionally, computational costs escalate exponentially as the cluster size goes up. Through the analysis, a cluster size of $20$ to $50$ seems to strike the right balance between performance and computational efficiency. This observation not only validates the initial rationale for clustering users, but also underscores the significance of computational efficiency in optimizing social campaigns under budget constraints.

\noindent \emph{Boosted Exploration Scores:}
Bandit algorithms aim to strike a delicate balance between exploiting known information to maximize short-term gains and exploring unknown options to improve long-term performance. In this spirit, we also consider in the experiments a variant of exploration, in which we
\emph{boost} the exploration score of unchosen arms having zero reward outcomes, to increase the likelihood of exploring alternative arms. We run experiments on the Twitter dataset with $L=2$ comparing the use of such artificially boosted exploration against its absence. The results are presented in Fig.~\ref{fig: arti_explore} in  the appendix, and  confirm the effectiveness of this approach;  this further supports  the importance of exploration to uncover valuable insights and optimize online decision making. 

\vspace{-1mm}
\section{Conclusion}
\vspace{-1mm}
In summary, our IM-GNB framework seamlessly leverages the expressivity of GNBs to tackle the  challenges of multi-round  IM in uncertain environments. Our novel approach tackles key issues in learning from graph-structured data and makes sequential decisions in uncertain environments. By incorporating contextual bandits, we obtain initial estimates of diffusion probabilities to construct exploitation and exploration graphs. Subsequently, these estimates are refined with GCNs, to enhance the influence spread. The framework's scalability, even without prior knowledge of the network topology, makes it a valuable and versatile tool for optimizing diffusion campaigns.

 \subsubsection*{Acknowledgements} This  work is funded by the Singapore Ministry of Education AcRF Tier 2 (A-8000423-00-00). This research is part of the programme DesCartes and is supported by the National Research Foundation, Prime Minister's Office, Singapore under its Campus for Research Excellence and Technological Enterprise (CREATE) programme. The authors  thank Fengzhuo Zhang and Junwen Yang (both   NUS) for valuable discussions.

\newpage
\bibliographystyle{ACM-Reference-Format}
\bibliography{ref}


\appendix


\newpage

\section{Supplementary complexity experiments}

We provide in Table \ref{tab:tab_datasets_orig1}  a comparison on   running time (in hours) w.r.t.\ the number of clustering groups $m'$ when $L=2$. We can observe that a finer granularity (more groups) may not result in better performance (as shown in Fig.~\ref{fig:group}), while cost goes up exponentially. We chose 50 groups that represents a good tradeoff between accuracy and complexity in our experiments.

\begin{table}[h]
  \caption{Results for running time.}
  \label{tab:tab_datasets_orig1}
  \scalebox{0.76}
  {
  \begin{tabular}{lccccccccc}
    \toprule
         {$m'$} &2&5&10&20&50&100&150&200&250\\
    \midrule
          running time&5.35&7.02&9.71&12.35&41.43&125.49&225.67&391.54&735.35\\
  \bottomrule
\end{tabular}
}
\end{table}


\section{Analysis on the number of clustering groups and for artificial exploration}

\begin{figure}[h]
    \centering
\includegraphics[width=0.85\columnwidth]{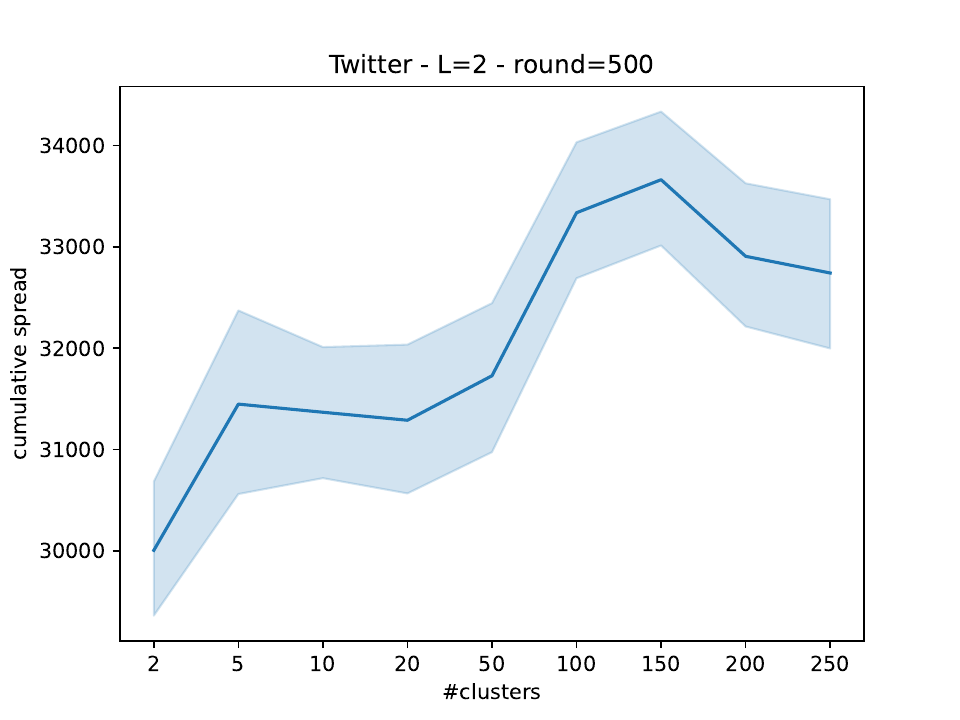}
\caption{Analysis on the number of clustering groups.} \label{fig:group} \vspace{.1in}
\includegraphics[width=0.85\columnwidth]{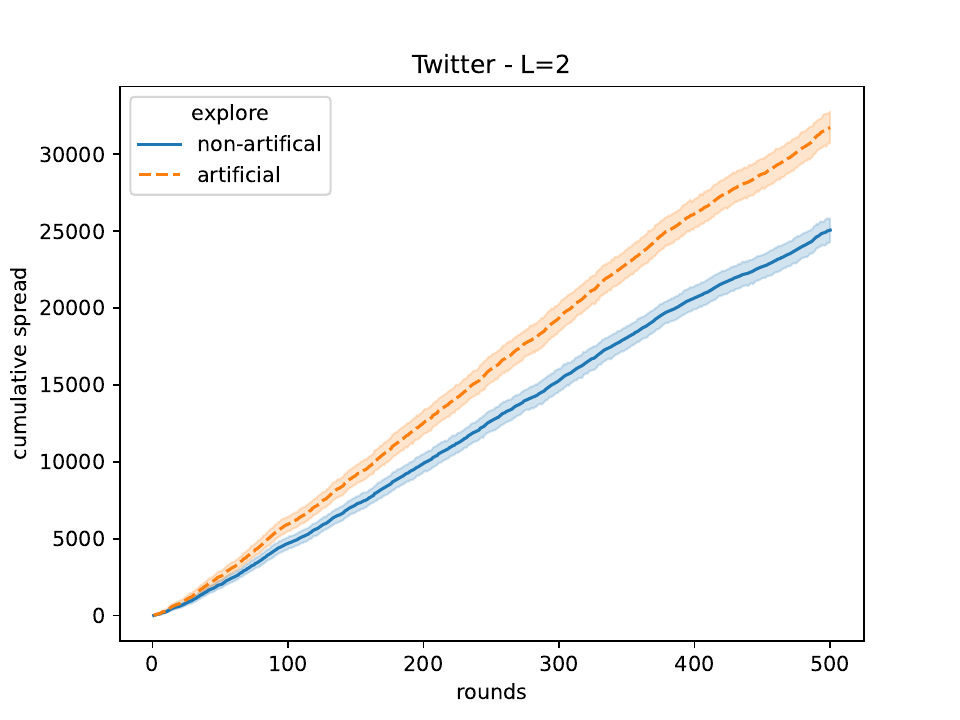}
\caption{Analysis on artificially boosted exploration.}
  \label{fig: arti_explore}
\end{figure}

Here, we present the remaining figures (Figs.~\ref{fig:group} and~\ref{fig: arti_explore}) that were mentioned in the main text but omitted due to space considerations. They pertain to the analysis of the number of clustering groups and for artificially boosted exploration.

Fig.~\ref{fig:group} shows that the campaign performance improves as the number of clusters increases from $2$ to $150$ at the beginning. However, beyond $200$ clusters, performance begins to decline. This decline can be likely attributed to insufficient data within each cluster for effective learning under the constraints of a limited rounds budget.

  Fig.~\ref{fig: arti_explore} confirms the effectiveness of artificially augmenting / boosting the exploration score of the unchosen arms with zero reward outcomes, in order to increase the likelihood of exploring alternative arms.  




\end{document}